\documentclass{article}
\usepackage{spconf,amsmath,graphicx}

\usepackage{booktabs}
\usepackage{adjustbox}
\usepackage{multirow}
\usepackage{textcomp}
\usepackage{caption}
\usepackage{subcaption}

\title{Identification of Indian Languages using Ghost-VLAD pooling}
\name{Krishna D N, Ankita Patil, M.S.P Raj,Sai Prasad H S, Prabhu Aashish Garapati}
\address{Youplus India, Bangalore\\
    \{krishna, raj,ankita,saiprasad,aashish\}@youplus.com}
\begin{document}
%
\maketitle
\begin{abstract}
In this work, we propose a new pooling strategy for language identification by considering Indian languages. The idea is to obtain utterance level features for any variable length audio for robust language recognition. We use the GhostVLAD approach to generate an utterance level feature vector for any variable length input audio by aggregating the local  frame level features across time. The generated feature vector is shown to have very good language discriminative features and helps in getting state of the art results for language identification task. We conduct our experiments on 635Hrs of audio data for 7 Indian languages. Our method outperforms the previous state of the art x-vector [11] method by an absolute improvement of 1.88\% in F1-score and achieves 98.43\% F1-score on the held-out test data. We compare our system with various pooling approaches and show that GhostVLAD is the best pooling approach for this task. We also provide visualization of the utterance level embeddings generated using Ghost-VLAD pooling and show that this method creates embeddings which has very good language discriminative features.

\end{abstract}
\section{INTRODUCTION}
\label{sec:intro}
The idea of language identification is to classify a given audio signal into a particular class using a classification algorithm. Commonly language identification task was done using i-vector systems [1]. A very well known approach for language identification proposed by N. Dahek et al. [1] uses the GMM-UBM model to obtain utterance level features called i-vectors. Recent advances in deep learning [15,16] have helped to improve the language identification task using many different neural network architectures which can be trained efficiently using GPUs for large scale datasets. These neural networks can be configured in various ways to obtain better accuracy for language identification task. Early work on using Deep learning for language Identification was published by Pavel Matejka et al. [2], where they used stacked bottleneck features extracted from deep neural networks for language identification task and showed that the bottleneck features learned by Deep neural networks are better than simple MFCC or PLP features. Later the work by I. Lopez-Moreno et al. [3] from Google showed how to use Deep neural networks to directly map the sequence of MFCC frames into its language class so that we can apply language identification at the frame level. Speech signals will have both spatial and temporal information, but simple DNNs are not able to capture temporal information. Work done by J. Gonzalez-Dominguez et al. [4] by Google developed an LSTM based language identification model which improves the accuracy over the DNN based models. Work done by Alicia et al. [5] used CNNs to improve upon i-vector [1] and other previously developed systems. The work done by Daniel Garcia-Romero et al. [6] has used a combination of Acoustic model trained for speech recognition with Time-delay neural networks where they train the TDNN model by feeding the stacked bottleneck features from acoustic model to predict the language labels at the frame level. Recently X-vectors [7] is proposed for speaker identification task and are shown to outperform all the previous state of the art speaker identification algorithms and are also used for language identification by David Snyder et al. [8].

In this paper, we explore multiple pooling strategies for language identification task. Mainly we propose Ghost-VLAD based pooling method for language identification. Inspired by the recent work by W. Xie et al. [9] and Y. Zhong et al. [10], we use Ghost-VLAD to improve the accuracy of language identification task for Indian languages. We explore multiple pooling strategies including NetVLAD pooling [11], Average pooling and Statistics pooling( as proposed in X-vectors [7]) and show that Ghost-VLAD pooling is the best pooling strategy for language identification. Our model obtains the best accuracy of 98.24\%, and it outperforms all the other previously proposed pooling methods. We conduct all our experiments on 635hrs of audio data for 7 Indian languages collected from $\boldsymbol{\textbf{All India Radio}}$ news channel\footnote{http://www.newsonair.com/}. The paper is organized as follows. In section 2, we explain the proposed pooling method for language identification. In section 3, we explain our dataset. In section 4, we describe the experiments, and in section 5, we describe the results.

\section{POOLING STRATEGIES}
In any language identification model, we want to obtain utterance level representation which has very good language discriminative features. These representations should be compact and should be easily separable by a linear classifier. The idea of any pooling strategy is to pool the frame-level representations into a single utterance level representation. Previous works by [7] have used simple mean and standard deviation aggregation to pool the frame-level features from the top layer of the neural network to obtain the utterance level features. Recently [9] used VLAD based pooling strategy for speaker identification which is inspired from [10] proposed for face recognition. The NetVLAD [11] and Ghost-VLAD [10] methods are proposed for Place recognition and face recognition, respectively, and in both cases, they try to aggregate the local descriptors into global features. In our case, the local descriptors are features extracted from ResNet [15], and the global utterance level feature is obtained by using GhostVLAD pooling.  In this section, we explain different pooling methods, including NetVLAD, Ghost-VLAD, Statistic pooling, and Average pooling.

\subsection{NetVLAD pooling }
\label{ssec:subhead}
The NetVLAD pooling strategy was initially developed for place recognition by R. Arandjelovic et al. [11]. The NetVLAD is an extension to VLAD [18] approach where they were able to replace the hard assignment based clustering with soft assignment based clustering so that it can be trained with neural network in an end to end fashion. In our case, we use the NetVLAD layer to map N local features of dimension D into a fixed dimensional vector, as shown in Figure 1 (Left side).

\begin{figure}[!htbp]
  \centering
\minipage{0.5\textwidth}
  \includegraphics[width=\linewidth]{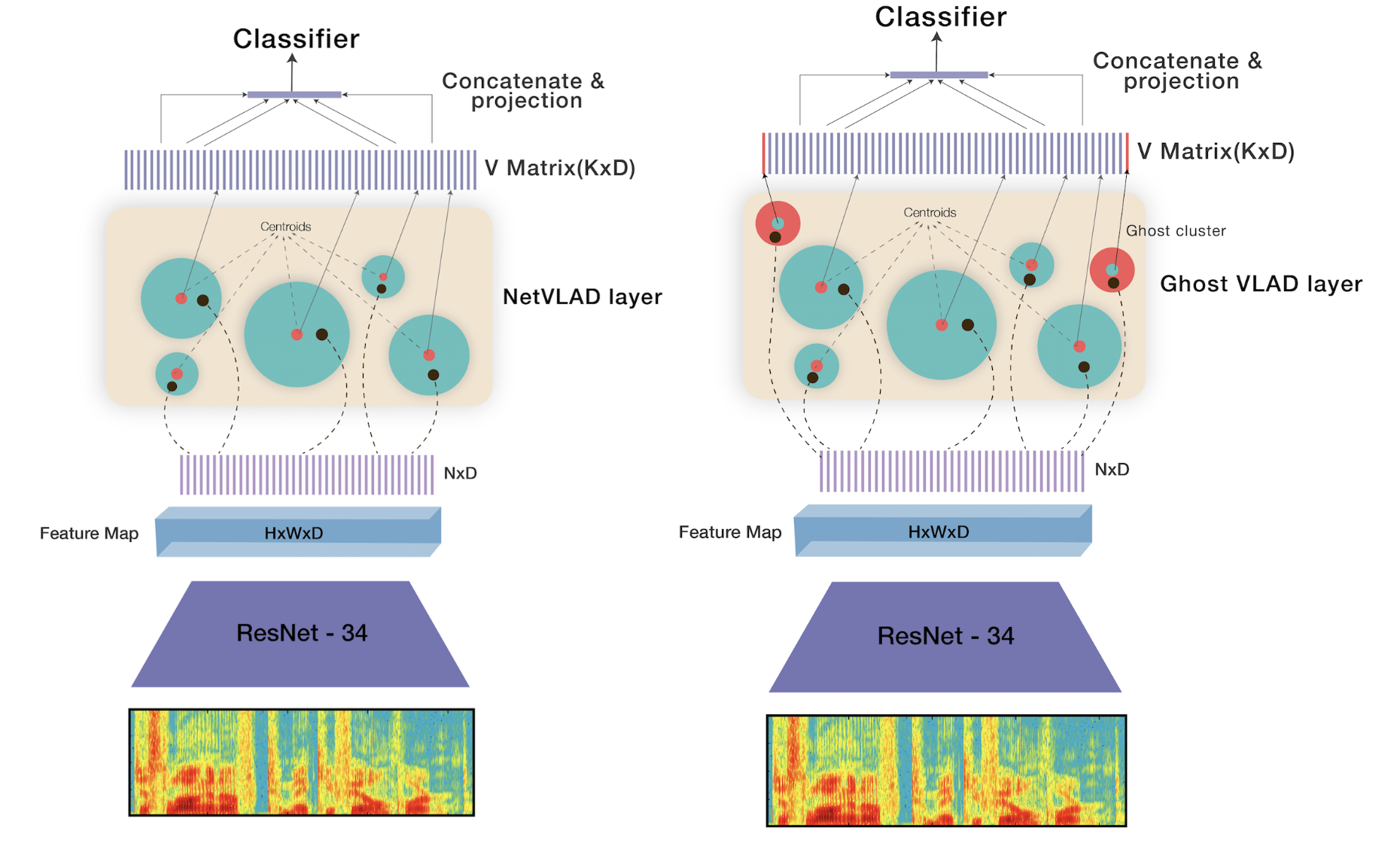}
  \caption{NetVLAD(Left side) and GhostVLAD(Right side)}
    \label{fig:tree-net}
  \endminipage\hfill
\end{figure}

The model takes spectrogram as an input and feeds into CNN based ResNet architecture. The ResNet is used to map the spectrogram into 3D feature map of dimension HxWxD. We convert this  3D feature map into 2D by unfolding H and W dimensions, creating a NxD dimensional feature map, where N=HxW. The NetVLAD layer is kept on top of the feature extraction layer of ResNet, as shown in Figure 1. The NetVLAD now takes N features vectors of dimension D and computes a matrix V of dimension KxD, where K is the number clusters in the NetVLAD layer, and D is the dimension of the feature vector. The matrix V is computed as follows.

\begin{equation}
{V(j,k)} =  {\sum_{i=1}^{N}\frac{{e}^{\boldsymbol{w}_k^{T}\boldsymbol{x_i}+b_k}}{\sum_{k'}^{K}{e}^{\boldsymbol{w}_k'^{T}\boldsymbol{x_i}+b_k'}}} (x_i(j) - c_k(j))
\end{equation}
Where $\boldsymbol{w_k}$,$\boldsymbol{b_k}$ and $\boldsymbol{c_k}$ are trainable parameters for the cluster $\boldsymbol{k}$  and V(j,k) represents a point in the V matrix for (j,k)th location. The matrix is constructed using the equation (1) where the first term corresponds to the soft assignment of the input $\boldsymbol{x_i}$ to the cluster $\boldsymbol{c_k}$, whereas the second term corresponds to the residual term which tells how far the input descriptor $\boldsymbol{x_i}$ is from the cluster center $\boldsymbol{c_k}$.

\subsection{GhostVLAD pooling }
\label{ssec:subhead}
GhostVLAD is an extension of the NetVLAD approach, which we discussed in the previous section. The GhostVLAD model was proposed for face recognition by Y. Zhong [10]. GhostVLAD works exactly similar to NetVLAD except it adds Ghost clusters along with the NetVLAD clusters. So, now we will have a K+G number of clusters instead of K clusters. Where G is the number of ghost clusters, we want to add (typically 2-4). The Ghost clusters are added to map any noisy or irrelevant content into ghost clusters and are not included during the feature aggregation stage, as shown in Figure 1 (Right side). Which means that we compute the matrix V for both normal cluster K and ghost clusters G, but we will not include the vectors belongs to ghost cluster from V during concatenation of the features. Due to which, during feature aggregation stage the contribution of the noisy and unwanted features to normal VLAD clusters are assigned less weights while Ghost clusters absorb most of the weight. We illustrate this in Figure 1(Right Side), where the ghost clusters are shown in red color. We use Ghost clusters when we are computing the V matrix, but they are excluded during the concatenation stage. These concatenated features are fed into the projection layer, followed by softmax to predict the language label.

\subsection{Statistic and average pooling }
\label{ssec:subhead}
In statistic pooling, we compute the first and second order statistics of the local features from the top layer of the ResNet model. The 3-D feature map is unfolded to create N features of D dimensions, and then we compute the mean and standard deviation of all these N vectors and get two D dimensional vectors, one for mean and the other for standard deviation. We then concatenate these 2 features and feed it to the projection layer for predicting the language label.

In the Average pooling layer, we compute only the first-order statistics (mean) of the local features from the top layer of the CNN model. The feature map from the top layer of CNN  is unfolded to create N features of D dimensions, and then we compute the mean of all these N vectors and get D dimensional representation. We then feed this feature to the projection layer followed by softmax for predicting the language label.

\section{DATASET}
\label{sec:print}
In this section, we describe our dataset collection process. We collected and curated around 635Hrs of audio data for 7 Indian languages, namely Kannada, Hindi, Telugu, Malayalam, Bengali, and English. We collected the data from the All India Radio news channel where an actor will be reading news for about 5-10 mins. To cover many speakers for the dataset, we crawled data from 2010 to 2019. Since the audio is very long to train any deep neural network directly, we segment the audio clips into smaller chunks using Voice activity detector\footnote{https://github.com/wiseman/py-webrtcvad}. Since the audio clips will have music embedded during the news, we use Inhouse music detection model to remove the music segments from the dataset to make the dataset clean and our dataset contains 635Hrs of clean audio which is divided into ~520Hrs of training data containing ~165K utterances and ~115Hrs of testing data containing ~35K utterances. The amount of audio data for training and testing for each of the language is shown in the table bellow.

\begin{table}[!htbp]
  \centering
  \label{tab:tasks}
  \begin{adjustbox}{width=.35\textwidth}
    \begin{tabular}{lcc}
      \toprule
      \textbf{Language} & \textbf{Training (hrs)} & \textbf{Testing (hrs)}\\
      \midrule
      Hindi & 113.08 & 19.65\\
      English & 55.86 & 19.19\\
      Kannada & 104.47 & 18.94\\
      Telugu & 110.01 & 11.62\\
      Assamese & 32.92 & 14.29\\
      Bengali & 63.59 & 20.62\\
      Malayalam & 37.82 & 10.88\\
      \bottomrule
    \end{tabular}
  \end{adjustbox}
  \caption{Dataset}

\end{table}

\section{EXPERIMENTS}
\label{sec:page}
In this section, we describe the feature extraction process and network architecture in detail.
We use spectral features of 256 dimensions computed using 512 point FFT for every frame, and we add an energy feature for every frame giving us total 257 features for every frame. We use a window size of 25ms and frame shift of 10ms during feature computation. We crop random 5sec audio data from each utterance during training which results in a spectrogram of size 257x500 (features x number of features). We use these spectrograms as input to our CNN model during training. During testing, we compute the prediction score irrespective of the audio length.

For the network architecture, we use ResNet-34 architecture, as described in [9]. The model uses convolution layers with Relu activations to map the spectrogram of size 257x500 input into 3D feature map of size 1x32x512. This feature cube is converted into 2D feature map of dimension 32x512 and fed into Ghost-VLAD/NetVLAD layer to generate a representation that has more language discrimination capacity. We use Adam optimizer with an initial learning rate of 0.01 and a final learning rate of 0.00001 for training. Each model is trained for 15 epochs with early stopping criteria.

For the baseline, we train an i-vector  model using GMM-UBM. We fit a small classifier on top of the generated i-vectors to measure the accuracy. This model is referred as {\fontfamily{qcr}\selectfont i-vector+svm }. To compare our model with the previous state of the art system, we set up the x-vector language identification system [8]. The x-vector model used time-delay neural networks (TDNN) along with statistic-pooling. We use 7 layer TDNN architecture similar to [8] for training. We refer to this model as {\fontfamily{qcr}\selectfont tdnn+stat-pool }. Finally, we set up a Deep LSTM based language identification system similar to [4] but with little modification where we add statistics pooling for the last layers hidden activities before classification. We use 3 layer Bi-LSTM with 256 hidden units at each layer. We refer to this model as {\fontfamily{qcr}\selectfont LSTM+stat-pool}. We train our {\fontfamily{qcr}\selectfont i-vector+svm} and  {\fontfamily{qcr}\selectfont TDNN+stat-pool} using Kaldi toolkit\footnote{https://kaldi-asr.org/}. We train our NetVLAD and GhostVLAD experiments using Keras\footnote{https://keras.io/} by modifying the code given by [9] for language identification. We train the {\fontfamily{qcr}\selectfont LSTM+stat-pool} and the remaining experiments using Pytorch [14] toolkit, and we will opensource all the codes and data soon.

\section{RESULTS}
\label{sec:illust}
In this section, we compare the performance of our system with the recent state of the art language identification approaches. We also compare different pooling strategies and finally, compare the robustness of our system to the length of the input spectrogram during training. We visualize the embeddings generated by the GhostVLAD method and conclude that the GhostVLAD embeddings shows very good feature discrimination capabilities.

\subsection{Comparison with different approaches }
\label{ssec:subhead}
We compare our system performance with the previous state of the art language identification approaches, as shown in Table 2. The  {\fontfamily{qcr}\selectfont i-vector+svm}  system is trained using GMM-UBM models to generate i-vectors as proposed in [1]. Once the i-vectors are extracted, we fit SVM classifier to classify the audio. The {\fontfamily{qcr}\selectfont TDNN+stat-pool} system is trained with a statistics pooling layer and is called the x-vector system as proposed by David Snyder et al. [11] and is currently the state of the art language identification approach as far as our knowledge. Our methods outperform the state of the art x-vector system by absolute 1.88\% improvement in F1-score, as shown in Table 2.
\begin{table}[!htbp]
  \centering
  \label{tab:tasks}
  \begin{adjustbox}{width=.35\textwidth}
    \begin{tabular}{lcc}
      \toprule
      \textbf{Methods} & \textbf{F1-Score (\%)} \\
      \midrule
      {\fontfamily{qcr}\selectfont i-vector+svm} & -\\
      {\fontfamily{qcr}\selectfont TDNN+stat-pool} & 96.52\\
      {\fontfamily{qcr}\selectfont LSTM+stat-pool} & 96.55\\
      {\fontfamily{qcr}\selectfont GhostVLAD} (ours)& 98.43\\
      \bottomrule
    \end{tabular}
  \end{adjustbox}
  \caption{Comparison Previous methods}
\end{table}
\subsection{Comparison with different pooling techniques }
\label{ssec:subhead}
We compare our approach with different pooling strategies in Table 3. We use ResNet as our base feature extraction network. We keep the base network the same and change only the pooling layers to see which pooling approach performs better for language identification task. Our experiments show that GhostVLAD pooling outperforms all the other pooling methods by achieving 98.43\% F1-Score.
\begin{table}[!htbp]
  \centering
  \label{tab:tasks}
  \begin{adjustbox}{width=.35\textwidth}
    \begin{tabular}{lcc}
      \toprule
      \textbf{Pooling Method} & \textbf{F1-Score (\%)} \\
      \midrule
       Average pooling & 97.23\\
       Statistics pooling & 97.38\\
       NetVLAD pooling & 98.13\\
       GhostVLAD pooling & 98.43\\
      \bottomrule
    \end{tabular}
  \end{adjustbox}
   \caption{Comparison with different Pooling methods }
\end{table}
\subsection{Duration analysis }
\label{ssec:subhead}
To observe the performance of our method with different input durations, we conducted an experiment where we train our model on different input durations. Since our model uses ResNet as the base feature extractor, we need to feed fixed-length spectrogram. We conducted 4 different experiments where we trained the model using 2sec, 3sec, 4sec and 5sec spectrograms containing 200,300,400 and 500 frames respectively. We observed that the model trained with a 5sec spectrogram is the best model, as shown in Table 4.

\begin{table}[!htbp]
  \centering
  \label{tab:tasks}
  \begin{adjustbox}{width=.35\textwidth}
    \begin{tabular}{lcc}
      \toprule
      \textbf{No. input frames (training) } & \textbf{F1-Score (\%)} \\
      \midrule
       200  & 98.18\\
       300  & 98.34\\
       400  & 98.28\\
       500  & 98.43\\
      \bottomrule
    \end{tabular}
  \end{adjustbox}
  \caption{F1-scores for different input sizes of the spectrogram}
\end{table}

\subsection{Visualization of embeddings}
\label{ssec:subhead}
We visualize the embeddings generated by our approach to see the effectiveness. We extracted 512-dimensional embeddings for our testing data and reduced the dimensionality using t-sne projection. The t-sne plot of the embeddings space is shown in Figure 3. The plot shows that the embeddings learned by our approach has very good discriminative properties

\begin{figure}[!htbp]
  \centering
\minipage{0.5\textwidth}
  \includegraphics[width=\linewidth]{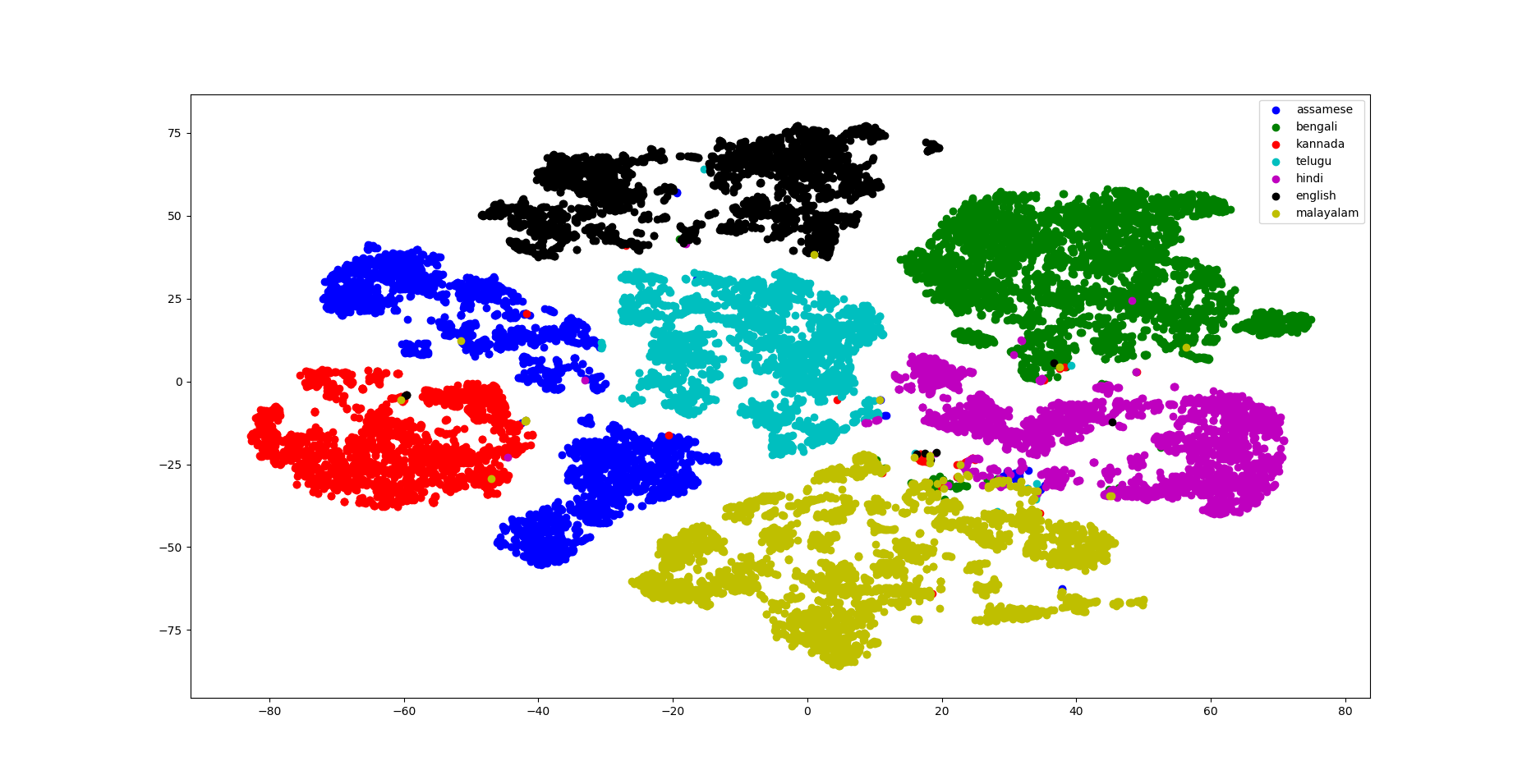}
  \caption{t-sne plot of embeddings}
    \label{fig:tree-net}
  \endminipage\hfill
\end{figure}

\section{Conclusion}
\label{sec:foot}

In this work, we use Ghost-VLAD pooling approach that was originally proposed for face recognition to improve language identification performance for Indian languages. We collected and curated ~630 hrs audio data from news All India Radio news channel for 7 Indian languages. Our experimental results shows that our approach outperforms the previous state of the art methods by an absolute 1.88\% F1-score. We have also conducted experiments with different pooling strategies proposed in the past, and the GhostVLAD pooling approach turns out to be the best approach for aggregating frame-level features into a single utterance level feature. Our experiments also prove that our approach works much better even if the input during training contains smaller durations. Finally, we see that the embeddings generated by our method has very good language discriminative features and helps to improve the performance of language identification.

\end{document}